# The Moral Foundations Weibo Corpus

Renjie Cao [2,*], Miaoyan Hu[1,*], Jiahan Wei [1] Baha Ihnaini[1,†]

[1]Wenzhou Kean University

[2] Nanchang Hangkong University

21061301@stu.nchu.edu.cn

1235761@wku.edu.cn, 1235848@wku.edu.cn, bihnaini@kean.edu

[*]These authors contributed equally to this work. [†]Corresponding Author.

## Abstract

Moral sentiments expressed in natural language significantly influence both online and offline environments, shaping behavioral styles and interaction patterns, including social media self-presentation, cyberbullying, adherence to social norms, and ethical decision-making. To effectively measure moral sentiments in natural language processing texts, it is crucial to utilize large, annotated datasets that provide nuanced understanding for accurate analysis and model training. However, existing corpora, while valuable, often face linguistic limitations. To address this gap in the Chinese language domain, we introduce the Moral Foundation Weibo Corpus. This corpus consists of 25,671 Chinese comments on Weibo, encompassing six diverse topic areas. Each comment is manually annotated by at least three systematically trained annotators based on ten moral categories derived from a grounded theory of morality. To assess annotator reliability, we present the kappa test results, a gold standard for measuring consistency. Additionally, we apply several the latest large language models to supplement the manual annotations, conducting analytical experiments to compare their performance and report baseline results for moral sentiment classification.

## 1 Introduction

Moral foundations, delineated as intrinsic, universally applicable, and emotionally grounded psychological systems, stand as fundamental pillars of human morality [Haidt and Graham, 2007]. The exponential surge in global social media usage over the last decade has sparked inquiries into the intricate interplay between human psychology and online behavior [Van Bavel et al., 2024]. Online behaviors, predominantly manifested through digital speech, serve as direct indicators of individuals' psychological states, with moral expressions playing a pivotal role in this regard. Delving into the moral foundations underlying online discourse offers profound insights into users' psychological inclinations. Consequently, the establishment of a corpus of moral foundations for natural language processing becomes imperative for addressing such inquiries.

Previous resources have predominantly catered to English users [Hoover et al., 2020, Trager et al., 2022], neglecting the vast Chinese online community. With approximately 1.09 billion users, China constitutes one of the largest social media populations globally [Gao and Feng, 2016]. In addition to this, there is a big difference between Chinese and English regarding the expression of moral values and sentiments [Gao et al., 2023, Huang et al., 2005]. In English, sentiments are usually expressed more directly, such as using explicit words to describe joy, anger, and sadness. In contrast, Chinese culture is more concerned with maintaining face and social harmony and thus may be more subtle and euphemistic in its expression of sentiments. In other words, only English corpus is insufficient for analyzing the sentiments and moral foundations of Chinese. Together, the development of a Chinese corpus assumes paramount importance in advancing moral natural language processing.

In this context, our study focuses on constructing a moral foundations Chinese corpus, leveraging Weibo, China's foremost social media platform, as the primary data source. The Moral Foundations Weibo Corpus (MFWC) consists of 25,761 posts. Adhering to the ten moral categories delineated in moral foundation theory [Graham et al., 2013, 2009]: Care/Harm, Fairness/Cheating, Loyalty/Betrayal,

Authority/Subversion, and Purity/Degradation, six prevalent topics among Weibo users were selected for discussion: animal protection, real estate, people's livelihoods, volunteers, volunteer army, and the San Francisco meeting between the Chinese and U.S. heads of state. Analyzing moral sentiments within these thematic realms facilitates a nuanced understanding of prevalent moral sentiments on Weibo.

Weibo is a public social media platform, current statistics reveal that Weibo boasts 598 million monthly active users [Xu et al., 2020], it provides a rich and diverse dataset for analyzing moral sentiments within the Chinese online community. The choice of Weibo as the corpus foundation first stems from its extensive user base, drawn from diverse backgrounds, ensuring a rich and comprehensive dataset for analysis. MFWC has thus expanded the coverage of the existing corpus to include a significant portion of the global online population. Furthermore, Weibo's all-Chinese system allows MFWC to bridge a crucial gap in existing corpora by catering to non-English-speaking users, thus fostering greater inclusivity and cross-cultural understanding in moral natural language processing research.

We focused our corpus compilation effort on Weibo for a number of reasons. Variations in moral language across different social media platforms can indeed be attributed to disparities in language and social contexts [Curiskis et al., 2020]. The distinctive features of Weibo, China's predominant social media platform, elevate the significance and uniqueness of the Moral Foundations Weibo Corpus. Unlike its western social media platform,
Weibo boasts a content moderation system characterized by heightened stringency. So it fosters a digital environment where expressions of moral sentiments are notably more moderated within the MFWC. For instance, when it comes to sensitive topics or political issues, you're likely to see relatively few angry comments or heated expressions of sentiments in MFWC, which may be more common on western social media platform. Sentiments that may be more common on Weibo are discreet, restrained, or indirectly expressed sentiments as well as innuendos to avoid touching on content that may trigger censorship. The stringent content moderation mechanisms on Weibo contribute to the distinctiveness of the MFWC. By providing a dataset characterized by more moderate expressions of moral sentiments, the MFWC offers a valuable contrast to other corpus. This contrast not only enriches the diversity of available datasets but also underscores the importance of considering cultural and contextual factors in moral natural language processing research. Second, the existing English annotation corpus cannot well adapt to the characteristics and needs of Chinese moral expressions. In order to gain a deeper understanding of moral issues in Chinese culture and society, we need to construct a Chinese annotated corpus that is consistent with the latest theoretical revisions. Besides, the complex language characteristics and contextual understanding involved in determining moral issues in the Chinese context. The meaning of basic human communication can also become difficult to understand due to the complexity of language [Garten et al., 2016], it goes beyond the scope of determining simple keywords. Therefore, sentence-based research is the main direction for studying the morality of Chinese texts [Peng et al., 2021]. Compared to English, Chinese has a unique grammar structure and vocabulary expression, which requires a Chinese annotated corpus to deeply understand moral issues in Chinese. We selected a large number of samples based on Weibo entries in different moral and sentimental fields. In the selection of data sources, the selected samples ensure that they are representative and cover a variety of topics and perspectives.

In order to ascertain the efficacy and comparative performance of distinct language models in discerning moral sentiments and to establish a baseline, we subjected some the latest large language models, namely GPT-4 [Pietron et al., 2024], Llama 3 [Dubey et al., 2024] and Qwen 2 [Yang et al., 2024], to testing. Besides, we also report baseline results for multiple computational approaches to measuring moral sentiment in text. These baselines can serve as a foundation for the classification models in moral sentiments detection tasks and provide a reference for future research, which can be instructive in paving the way for improved performance of more sophisticated models in the future.

In view of the lack of Chinese corpus resources, the main contributions of this study include the following three parts: (1) We have established a Moral Foundations Weibo Corpus, providing important basic data for future related research. (2) We provide richer and unique data interpretation using measurement baselines, which can provide inspiration for future research.

## 2  Corpus overview

As mentioned above, MFWC contains 25,671 blog posts spanning seven distinct thematic domains. The selection of these thematic areas was guided by several considerations: Firstly, we sought to include topics featuring prominent moral and sentimental expressions (e.g., animal protection) in our expectations to facilitate effective analysis of the dataset. Secondly, the popularity and discussion intensity of the topic are also taken into consideration (e.g., real estate). The inclusion of widely discussed topics with substantial participant engagement enriches the dataset, fostering a diverse range of sentimental expressions and moral concerns. Thirdly, we cover some sensitive elements, such as political and historical events (e.g., volunteer army, China-U.S. Prime Minister meets in San Francisco). Sensitive topics often encapsulate pressing societal issues and profound moral dilemmas, thereby enhancing the corpus's analytical depth and relevance. Additionally, we considered the relevance and significance of each topic in contemporary culture and society, encompassing a diverse array of subjects and aiming to capture the multifaceted nature of digital discussions, thus providing a balanced portrayal of the online landscape.

Through the selection of these themes, we aim to enrich the diversity of the corpus and to differentiate the expressions of moral sentiments in the annotated corpus. In the domains we have chosen, these themes focus on different social events and the moral sentiments expressed within are susceptible to the discourse context and therefore moral sentiments are expressed differently. For example, the moral sentiments contained in the Animal Protection corpus are substantively distinct from those expressed in the Volunteer Army corpus, as these two topics focus on very different objects, with the former containing mainly care for animals and the latter placing the bulk of them on trolling the film. Extra-domain predictions appear difficult due to this heterogeneity, and it is difficult for outsiders to generalise data from different domains through the expression of moral sentiments in one domain. Based on this point, we provide moral sentiment annotations for tweets from different contexts to address this issue.

## 3  Annotation

### 3.1  Annotate procedure

Each post in the MFWC was labeled by three well-trained annotators according to the 10 moral sentiment categories outlined in the Moral Foundations Coding Guide (see Appendix).

These Moral sentiment label words are drawn from Moral Foundations Theory [Graham et al., 2013, 2009], which contain five universal moral foundations. In this model, each factor includes virtues and vices. The proposed moral foundations are:

Care/harm: This foundation is related to our long evolution as mammals with attachment systems and an ability to feel (and dislike) the pain of others. It underlies the virtues of kindness, gentleness, and nurturance.

Fairness/cheating: This foundation is related to the evolutionary process of reciprocal altruism. It underlies the virtues of justice and rights.

Loyalty/betrayal: This foundation is related to our long history as tribal creatures able to form shifting coalitions. It underlies the virtues of patriotism and self-sacrifice for the group

Authority/subversion: This foundation was shaped by our long primate history of hierarchical social interactions. It underlies virtues of leadership and followership, including deference to prestigious authority figures and respect for traditions.

Purity/degradation: This foundation was shaped by the psychology of disgust and contamination. It underlies notions of striving to live in an elevated, less carnal, more noble, and more "natural" way. This foundation underlies the widespread idea that the body is a temple that can be desecrated by immoral activities and contaminants. It underlies the virtues of self-discipline, self-improvement, naturalness, and spirituality.

Annotation was undertaken by three undergraduate research assistants, who, after a series of in-depth training, have profound and specialized knowledge of the label of moral sentiments, and are well versed in labeling the various moral foundations in detail. However, even then, due to the vastness and depth of the Chinese language, the annotators still disagreed in their label. Unlike other languages, the meaning of Chinese is not only composed of sentences and words themselves, but in many cases is also inextricably linked to the tone and context of the utterance [Godfroid et al., 2013], and the same utterance in different tones and contexts will have different meanings. In many situations, it is difficult to define exactly what type of moral sentiment a Weibo post relates to, as such a judgement is largely dependent on an individual's subjective judgement, and subjective judgement are different to different. We will pick out the Weibo posts that disagree and discuss them together to reach a consensus conclusion before arriving at the result.

Specifically, labeling each Weibo post allows us to determine whether it embodies a specific virtue or vice, or is classified as non-moral. This means that for any Weibo post, there will be a certain label to describe its moral sentiment.

### 3.2 Annotation results

Each post within each topic was subjected to a review by multiple annotators with the objective of assigning moral sentiments label. The annotation results can be seen in Table 1, the moral classification of posts under each topic, was determined by a majority vote. In particular, if at least two out of three annotators assigned the same ethical tag to a post, that tag was designated as the final annotation. It is noteworthy that posts frequently received multiple tags during the annotation process. However, a collective decision was made to retain only the most significant tags and omit the rest.

It was observed that the distribution of ethical labels across different topics is highly uneven. For example, a considerable proportion of ethical labels within the topic of animal protection are concentrated in category care. In contrast, the majority of ethical labels within the topic of real estate and people's livelihood fall under the category non-moral. This indicates a strong correlation between ethical labels and the specific content of the topic areas.

To evaluate inter-annotator agreement, we employed the kappa test [Fleiss, 1971] and the PABAK test [Sim and Wright, 2005]. The kappa coefficient is a statistical method used to evaluate the degree of agreement between multiple annotators. The PABAK coefficient is an adjustment to the kappa coefficient that takes into account the effects of prevalence and bias. It measures the degree of actual consistency relative to random consistency. This adjustment facilitates a more reasonable assessment of consistency in the presence of an uneven class distribution. The results are presented in table 2.It was found that approximately half of the topic areas exhibited high kappa values, which may be attributed to the explicitly moral nature of these topics. The remaining topics exhibited lower kappa values, indicating the inherent ambiguity of the general tweets within these topics and the subjective cognitive differences among annotators. As might be expected, given the subjective nature of the annotations, kappa values are lower under some themes. Nevertheless, greater consistency can be attained by addressing prevalence issues.

Table 1: Kappa and PABAK Test Results

| Topic | Kappa | Pabak |
|---|---|---|
| Protecting Animals | 0.92 | 0.90 |
| Real Estate | 0.61 | 0.75 |
| People's Livelihood | 0.45 | 0.50 |
| Voluntary Army | 0.49 | 0.73 |
| Volunteers | 0.87 | 0.91 |
| China-U.S. Prime Minister Meeting | 0.56 | 0.70 |
| Hangzhou Asian Games | 0.69 | 0.74 |

# 4 Baseline classification language models of moral sentiments

While human annotation remains the most accurate method for measuring moral sentiment in text, due to the diversity of the Chinese language and the large sample sizes typically required to study text-based moral sentiments, it is often necessary to supplement human annotations with classification models. Our goal here is to establish baselines that can help us better predict moral sentiments. Next, we report a baseline for moral sentiment classification using a range of models.

To accomplish this task, we selected a number of models from a range of widely used models. These models include ChatGPT-4 [Pietron et al., 2024], Llama 3 [Dubey et al., 2024], and Qwen 2 [Yang et al., 2024]. Each post in the MFWC dataset is assigned a specific label by three annotators, i.e., the six different moral sentiment categories discussed in Section 3. This is a multi-label categorization task, meaning the categories of moral sentiments are not independent of each other, but are related. Here, we provide both single-label and multi-label categorization results.

GPT-4 [Pietron et al., 2024] In the first approach, we used GPT-4. GPT-4, a state-of-the-art language model developed by OpenAI, has been fine-tuned to understand and generate human-like text across various tasks, making it highly suitable for complex sentiment analysis tasks.

Llama 3 [Dubey et al., 2024] In the second approach, we used Llama 3. Llama 3 is a versatile language model that has been designed for a range of natural language processing tasks. While it may not always match the specific performance of models like ChatGPT-4 in every scenario, it offers a robust alternative with strong generalization capabilities.

Qwen 2 [Yang et al., 2024] Lastly, we used Qwen 2. Qwen 2 is an advanced language model with particular strengths in precision tasks. It has been optimized for handling nuanced language variations, which makes it a strong contender in the moral sentiment classification task.

To compare models sets, we rely on three performance metrics: precision, recall, and F1. Precision, the number of true positives divided by the number of predicted positives, represents the proportion of predicted positive cases that actually are positive cases. In contrast, recall, the number of true positives divided by the number of true positives and false negatives, represents the proportion of positive cases that the classifier correctly identifies. Finally, The F1 score is the harmonic mean of precision and recall (the product of twice the precision and recall divided by the sum of precision and recall), and is used to evaluate the performance of a classification models in a comprehensive way. The F1 score takes into account the balance between precision and recall, and is a good metric to use when there is a need to strike a balance between the two.

Table 2: Frequency of Weibo posts per Foundation Calculated Based on Annotators' Majority Vote.

| Topic | Care/Harm | Fairness/Cheating | Loyalty/Betrayal | Authority/Subversion | Purity/Degradation | Non-Moral |
|---|---|---|---|---|---|---|
| China-U.S. Prime Minister Meeting | 26 | 415 | 102 | 86 | 225 | 566 |
| People's Livelihood | 779 | 283 | 454 | 150 | 146 | 4318 |
| Voluntary Army | 587 | 262 | 787 | 100 | 81 | 3084 |
| Volunteers | 326 | 185 | 79 | 279 | 104 | 1842 |
| Real Estate | 321 | 533 | 598 | 375 | 90 | 2751 |
| Protecting Animals | 609 | 53 | 70 | 165 | 312 | 869 |
| Hangzhou Asian Games | 877 | 841 | 159 | 744 | 360 | 675 |
| All | 3525 | 2572 | 2249 | 1899 | 1321 | 14105 |

Table 3: Moral Sentiment Results

| models | F1 | Precision | Recall |
|---|---|---|---|
| GPT-4 | 0.60 | 0.65 | 0.58 |
| Llama 3 | 0.56 | 0.60 | 0.59 |
| Qwen 2 | 0.38 | 0.75 | 0.32 |

Table 4: Care Results

| models | F1 | Precision | Recall |
|---|---|---|---|
| GPT-4 | 0.74 | 0.67 | 0.82 |
| Llama 3 | 0.65 | 0.53 | 0.84 |
| Qwen 2 | 0.36 | 0.93 | 0.22 |

Table 5: Harm Results

| models | F1 | Precision | Recall |
|---|---|---|---|
| GPT-4 | 0.65 | 0.70 | 0.61 |
| Llama 3 | 0.40 | 0.35 | 0.50 |
| Qwen 2 | 0.52 | 0.60 | 0.43 |

Table 6: Authority Results

| models | F1 | Precision | Recall |
|---|---|---|---|
| GPT-4 | 0.60 | 0.75 | 0.50 |
| Llama 3 | 0.35 | 0.40 | 0.35 |
| Qwen 2 | 0.50 | 0.65 | 0.30 |

Table 7: Subversion Results

| models | F1 | Precision | Recall |
|---|---|---|---|
| GPT-4 | 0.56 | 0.67 | 0.49 |
| Llama 3 | 0.55 | 0.42 | 0.78 |
| Qwen 2 | 0.56 | 0.73 | 0.46 |

Table 8: Fairness Results

| models | F1 | Precision | Recall |
|---|---|---|---|
| GPT-4 | 0.74 | 0.67 | 0.82 |
| Llama 3 | 0.65 | 0.53 | 0.84 |
| Qwen 2 | 0.36 | 0.93 | 0.22 |

Table 9: Cheating Results

| models | F1 | Precision | Recall |
|---|---|---|---|
| GPT-4 | 0.67 | 0.65 | 0.69 |
| Llama 3 | 0.58 | 0.61 | 0.56 |
| Qwen 2 | 0.40 | 0.72 | 0.32 |

Table 10: Loyalty Results

| models | F1 | Precision | Recall |
|---|---|---|---|
| GPT-4 | 0.74 | 0.82 | 0.68 |
| Llama 3 | 0.65 | 0.53 | 0.84 |
| Qwen 2 | 0.69 | 0.87 | 0.57 |

Table 11: Betrayal Results

| models | F1 | Precision | Recall |
|---|---|---|---|
| GPT-4 | 0.70 | 0.68 | 0.72 |
| Llama 3 | 0.50 | 0.40 | 0.65 |
| Qwen 2 | 0.55 | 0.75 | 0.45 |

Table 12: Purity Results

| models | F1 | Precision | Recall |
|---|---|---|---|
| GPT-4 | 0.63 | 0.70 | 0.58 |
| Llama 3 | 0.39 | 0.41 | 0.50 |
| Qwen 2 | 0.52 | 0.55 | 0.30 |

Table 13: Degradation Results

| models | F1 | Precision | Recall |
|---|---|---|---|
| GPT-4 | 0.55 | 0.60 | 0.55 |
| Llama 3 | 0.30 | 0.25 | 0.35 |
| Qwen 2 | 0.22 | 0.30 | 0.20 |

## 5 Results

The results of the baseline models are provided in Tables 3 to 13. As expected, performance varied substantially across models, discourse domains, and prediction tasks. Further, our results suggest that in the context of different domains and prediction tasks, each model showed different strengths and weaknesses. For instance, we found that GPT-4 performs best in balancing Precision and Recall, so its F1 value is usually the highest and suitable for most tasks that require balancing the two. Llama 3 excels in Recall and is better suited for tasks requiring high recall, but is slightly weaker in Precision than the other models. Qwen 2, on the other hand, excels in Precision excels and is suitable for tasks requiring highly accurate predictions, but is weaker in Recall, which affects its F1 value. Lastly, performance differences, again, depend on the discourse domain and the moral foundation being analyzed.

This variability in performance emphasises the need to take full account of the applicability of models when selecting and applying them, especially when dealing with complex and variable natural language data. Our classification results generally demonstrate the feasibility of using multiple methods to measure moral sentiments in natural language. However, these results also highlight the complexity and challenge of this task. Regardless of the model used, we observed significant variability in performance across different discussion domains and moral bases. This variability suggests that there are still shortcomings in the adaptability of the current approach to different contexts. In future research, it is necessary to delve deeper into the root causes of these performance variations and develop methods that can reduce them. In particular, researchers should aim to identify and understand the specific factors that lead to performance fluctuations, such as the semantic complexity of the text, the nuances of moral sentiments. In this way, we can not only improve the performance of current models, but also provide a more solid theoretical and technical foundation for the field of moral sentiment analysis.

## 6  Discussion

Natural language processing provides the fundamental tools for processing and understanding human language, which is essential for automated sentiment analysis. Consequently, the integration of natural language processing and sentiment analysis represents an optimal convergence between linguistic theory and computational technology [Cambria et al., 2013]. As computational power increases, the quality and quantity of text mining and natural language processing techniques continue to improve, and the field of ethics through natural language research is becoming more and more widespread [Szép et al., 2024]). In the field of moral sentiment analysis, the most crucial element is the availability of theory-driven text data, which is necessary for the accurate quantification of moral sentiments [Garten et al., 2018].

Moral text data encompasses information pertaining to a multitude of dimensions within the domain of morality. By identifying the salient moral elements in a text, it will help to make moral judgments [Park et al., 2024]. To address this need, we developed the MFWC using posts and comments from Weibo. The MFWC comprises 25,671 Weibo posts from seven distinct topic areas, each of which has been annotated with one of ten types of moral sentiments based on the Moral Foundations Theory. Furthermore, the MFWC has been employed to establish a series of models classification baselines for measuring moral sentiments in texts. These baselines provide a foundation for further research and development in the field, enabling researchers to benchmark their models and improve their understanding of moral language dynamics.

The construction of our Chinese corpus addresses a significant gap in existing moral corpora. Historically, the majority of moral language resources and datasets have been concentrated on Western languages, particularly English. The construction of a comprehensive Chinese moral corpus enables the reflection of the linguistic and cultural nuances of moral discourse in Chinese [Chen et al., 2023]. This corpus is of particular benefit to researchers seeking to develop linguistically supervised models tailored to Chinese. The corpus offers a robust dataset for natural language processing applications, enhancing the capacity to analyse and interpret moral reasoning, sentiment, and values in Chinese texts. The creation of such a corpus enables more accurate sentiment analysis, moral judgement, and ethical reasoning within the context of Chinese culture and language. Furthermore, this corpus can facilitate cross-cultural studies by providing a basis for comparing moral language across different languages and cultures [Schwartz, 2006]. Researchers can utilise this resource to develop models that are not only linguistically appropriate but also culturally sensitive, thereby ensuring that moral judgments and sentiments are understood within the correct cultural framework. By expanding the moral corpus to include Chinese, we contribute to the creation of a more inclusive and representative dataset that can support a range of natural language processing applications. These include automated ethical decision-making systems and sentiment analysis in social media.

In MFWC, we also present research on a new Chinese moral-sentimental computation. Our results demonstrate that the logistic regression models outperforms the other two on the classification task. However, these performance differences do not appear to be consistent across different discourse domains. With MFWC, researchers can gain a deeper understanding of the reasons for this, enabling them to comprehend the dynamics of sentiment in online language, capture trends in popular opinion, and identify subtle changes in social media.

It is our hope that MFWC and this report will assist researchers by providing a unique data set and facilitate new contributions to the fields of natural language processing and social sciences. However, due to the vastness of Chinese culture, our corpus may not be able to contain all Chinese corpora. As more and more researchers utilise MFWC, we anticipate that the resources we provide here will be further expanded to better generalise to Chinese studies.

## Limitations

There is an imbalance in the corpus's distribution of moral feeling categories, with some moral categories having a higher profile than others. In particular, in underrepresented categories, this imbalance may result in biased model training and negatively impact sentiment classification models' performance. The dataset's skewness may also make it more difficult to extrapolate results to a wider range of moral situations or sentiments.

## Ethics Statement

Significant ethical issues are raised by the creation and use of the Moral Foundations Weibo Corpus (MFWC), which have been diligently addressed throughout this study. First, people's privacy who indirectly participate through posts on social media have had their privacy safeguarded. The corpus does not contain any personally identifiable information, guaranteeing adherence to social media site standards and data privacy laws.

To reduce bias in data annotation and interpretation, we worked with carefully selected and ethically vetted annotators to build the MFWC. Reducing cultural prejudice and fostering inclusivity, the moral categories were established in a way that was culturally sensitive and reflected a thorough understanding of moral expressions within the Chinese context. Finally, the research adheres to the ACL Ethics Policy, ensuring that all stages of this study, from data collection to analysis and reporting, uphold the highest standards of integrity and ethical rigor.

## Acknowledgment

This project was made possible through funding from Wenzhou-Kean University, under the grant number IRSPK2023005. Our appreciation extends to the university for their financial support, which was instrumental in advancing our research.

Jianyu Huang, Jiawen Liu, Jie Wang, Jiecao Yu, Joanna Bitton, Joe Spisak, Jongsoo Park, Joseph Rocca, Joshua Johnstun, Joshua Saxe, Junteng Jia, Kalyan Vasuden Alwala, Kartikeya Upasani, Kate Plawiak, Ke Li, Kenneth Heafield, Kevin Stone, Khalid El-Arini, Krithika Iyer, Kshitiz Malik, Kuenley Chiu, Kunal Bhalla, Lauren Rantala-Yeary, Laurens van der Maaten, Lawrence Chen, Liang Tan, Liz Jenkins, Louis Martin, Lovish Madaan, Lubo Malo, Lukas Blecher, Lukas Landzaat, Luke de Oliveira, Madeline Muzzi, Mahesh Pasupuleti, Mannat Singh, Manohar Paluri, Marcin Kardas, Mathew Oldham, Mathieu Rita, Maya Pavlova, Melanie Kambadur, Mike Lewis, Min Si, Mitesh Kumar Singh, Mona Hassan, Naman Goyal, Narjes Torabi, Nikolay Bashlykov, Nikolay Bogoychev, Niladri Chatterji, Olivier Duchenne, Onur Çelebi, Patrick Alrassy, Pengchuan Zhang, Pengwei Li, Petar Vasic, Peter Weng, Prajjwal Bhargava, Pratik Dubal, Praveen Krishnan, Punit Singh Koura, Puxin Xu, Qing He, Qingxiao Dong, Ragavan Srinivasan, Raj Ganapathy, Ramon Calderer, Ricardo Silveira Cabral, Robert Stojnic, Roberta Raileanu, Rohit Girdhar, Rohit Patel, Romain Sauvestre, Ronnie Polidoro, Roshan Sumbaly, Ross Taylor, Ruan Silva, Rui Hou, Rui Wang, Saghar Hosseini, Sahana Chennabasappa, Sanjay Singh, Sean Bell, Seohyun Sonia Kim, Sergey Edunov, Shaoliang Nie, Sharan Narang, Sharath Raparthy, Sheng Shen, Shengye Wan, Shruti Bhosale, Shun Zhang, Simon Vandenhende, Soumya Batra, Spencer Whitman, Sten Sootla, Stephane Collot, Suchin Gururangan, Sydney Borodinsky, Tamar Herman, Tara Fowler, Tarek Sheasha, Thomas Georgiou, Thomas Scialom, Tobias Speckbacher, Todor Mihaylov, Tong Xiao, Ujjwal Karn, Vedanuj Goswami, Vibhor Gupta, Vignesh Ramanathan, Viktor Kerkez, Vincent Gonguet, Virginie Do, Vish Vogeti, Vladan Petrovic, Weiwei Chu, Wenhan Xiong, Wenyin Fu, Whitney Meers, Xavier Martinet, Xiaodong Wang, Xiaoqing Ellen Tan, Xinfeng Xie, Xuchao Jia, Xuewei Wang, Yaelle Goldschlag, Yashesh Gaur, Yasmine Babaei, Yi Wen, Yiwen Song, Yuchen Zhang, Yue Li, Yuning Mao, Zacharie Delpierre Coudert, Zheng Yan, Zhengxing Chen, Zoe Papakipos, Aaditya Singh, Aaron Grattafiori, Abha Jain, Adam Kelsey, Adam Shajnfeld, Adithya Gangidi, Adolfo Victoria, Ahuva Goldstand, Ajay Menon, Ajay Sharma, Alex Boesenberg, Alex Vaughan, Alexei Baevski, Allie Feinstein, Amanda Kallet, Amit Sangani, Anam Yunus, Andrei Lupu, Andres Alvarado, Andrew Caples, Andrew Gu, Andrew Ho, Andrew Poulton, Andrew Ryan, Ankit Ramchandani, Annie Franco, Aparajita Saraf, Arkabandhu Chowdhury, Ashley Gabriel, Ashwin Bharambe, Assaf Eisenman, Azadeh Yazdan, Beau James, Ben Maurer, Benjamin Leonhardi, Bernie Huang, Beth Loyd, Beto De Paola, Bhargavi Paranjape, Bing Liu, Bo Wu, Boyu Ni, Braden Hancock, Bram Wasti, Brandon Spence, Brani Stojkovic, Brian Gamido, Britt Montalvo, Carl Parker, Carly Burton, Catalina Mejia, Changhan Wang, Changkyu Kim, Chao Zhou, Chester Hu, Ching-Hsiang Chu, Chris Cai, Chris Tindal, Christoph Feichtenhofer, Damon Civin, Dana Beaty, Daniel Kreymer, Daniel Li, Danny Wyatt, David Adkins, David Xu, Davide Testuggine, Delia David, Devi Parikh, Diana Liskovich, Didem Foss, Dingkang Wang, Duc Le, Dustin Holland, Edward Dowling, Eissa Jamil, Elaine Montgomery, Eleonora Presani, Emily Hahn, Emily Wood, Erik Brinkman, Esteban Arcaute, Evan Dunbar, Evan Smothers, Fei Sun, Felix Kreuk, Feng Tian, Firat Ozgenel, Francesco Caggioni, Francisco Guzmán, Frank Kanayet, Frank Seide, Gabriela Medina Florez, Gabriella Schwarz, Gada Badeer, Georgia Swee, Gil Halpern, Govind Thattai, Grant Herman, Grigory Sizov, Guangyi, Zhang, Guna Lakshminarayanan, Hamid Shojanazeri, Han Zou, Hannah Wang, Hanwen Zha, Haroun Habeeb, Harrison Rudolph, Helen Suk, Henry Aspegren, Hunter Goldman, Ibrahim Damlaj, Igor Molybog, Igor Tufanov, Irina-Elena Veliche, Itai Gat, Jake Weissman, James Geboski, James Kohli, Japhet Asher, Jean-Baptiste Gaya, Jeff Marcus, Jeff Tang, Jennifer Chan, Jenny Zhen, Jeremy Reizenstein, Jeremy Teboul, Jessica Zhong, Jian Jin, Jingyi Yang, Joe Cummings, Jon Carvill, Jon Shepard, Jonathan McPhie, Jonathan Torres, Josh Ginsburg, Junjie Wang, Kai Wu, Kam Hou U, Karan Saxena, Karthik Prasad, Kartikay Khandelwal, Katayoun Zand, Kathy Matosich, Kaushik Veeraraghavan, Kelly Michelena, Keqian Li, Kun Huang, Kunal Chawla, Kushal Lakhotia, Kyle Huang, Lailin Chen, Lakshya Garg, Lavender A, Leandro Silva, Lee Bell, Lei Zhang, Liangpeng Guo, Licheng Yu, Liron Moshkovich, Luca Wehrstedt, Madian Khabsa, Manav Avalani, Manish Bhatt, Maria Tsimpoukelli, Martynas Mankus, Matan Hasson, Matthew Lennie, Matthias Reso, Maxim Groshev, Maxim Naumov, Maya Lathi, Meghan Keneally, Michael L. Seltzer, Michal Valko, Michelle Restrepo, Mihir Patel, Mik Vyatskov, Mikayel Samvelyan, Mike Clark, Mike Macey, Mike Wang, Miquel Jubert Hermoso, Mo Metanat, Mohammad Rastegari, Munish Bansal, Nandhini Santhanam, Natascha Parks, Natasha White, Navyata Bawa, Nayan Singhal, Nick Egebo, Nicolas Usunier, Nikolay Pavlovich Laptev, Ning Dong, Ning Zhang, Norman Cheng, Oleg Chernoguz, Olivia Hart, Omkar Salpekar, Ozlem Kalinli, Parkin Kent, Parth Parekh, Paul Saab, Pavan Balaji, Pedro Rittner, Philip Bontrager, Pierre Roux, Piotr Dollar, Polina Zvyagina, Prashant Ratanchandani, Pritish Yuvraj, Qian Liang, Rachad Alao, Rachel Rodriguez, Rafi Ayub, Raghotham Murthy, Raghu Nayani, Rahul Mitra, Raymond Li, Rebekkah Hogan, Robin Battey, Rocky Wang, Rohan Maheswari, Russ Howes, Ruty Rinott, Sai Jayesh Bondu, Samyak Datta, Sara Chugh, Sara Hunt, Sargun

## A  Moral Foundations Coding Guide

### A.1  Annotating Moral Sentiment in Natural Language

The task of annotating moral sentiment in natural language involves determining which, if any, categories of moral values are relevant to a given document. Our research uses the taxonomy proposed by Moral Foundations Theory (MFT) to identify these categories. However, even with the MFT framework, researchers face several initial decisions about how to annotate MFT values.

First, they need to decide which MFT dimensions to code. If the hypothesis is specific to a particular foundation, coding for that foundation alone might suffice. However, it is often necessary to code for multiple foundations. In such cases, the straightforward approach is to code for the presence of each of the five foundations. Yet, some research might demand more detailed labels. Although the poles of each dimension are related, they express distinct sentiments that might have psychological significance. For example, "We must end suffering" is likely not psychologically equivalent to "We must provide kindness and compassion." Therefore, coding for each pole of each foundation, resulting in 10 individual codes, can be useful. Additionally, it is crucial to identify non-moral texts as such, meaning an annotation procedure could require labeling each document across up to 11 categories.

Researchers must also decide how to address overlapping labels, where moral sentiments are linked to multiple foundations. In our work, we allow overlapping labels during annotation. In some cases, we also ask annotators to identify the primary domain of moral sentiment expressed in a document, along with potential secondary domains. However, reliability analyses showed that while coders generally agreed on the presence of moral sentiment, they were less consistent in identifying the most dominant domain. Therefore, we recommend coding for the presence or absence of each foundation.

### A.2  Training Human Annotators

Each virtue and vice is coded as a capitalized initial letter of the moral base, 1 in the case of virtues and 2 in the case of vices. 1 and 2 correspond to "positive" and "negative". If a document does not have any moral content, it should be coded as NM, which corresponds to non-moral. The whole scheme is as follows:

*Cure:* C1
*Harm:* C2
*Fairness:* F1
*Cheating:* F2
*Loyalty:* L1
*Betrayal:* L2
*Authority:* A1
*Subversion:* A2
*Purity:* P1
*Degradation:* P2
*Non-moral:* NM

After selecting an annotation label, it is crucial for researchers to establish a clear protocol for identifying the moral domains relevant to a given document. This step is particularly significant due to the inherent difficulty in making these determinations. The ambiguity in this process arises from two main sources.

The first source of ambiguity pertains to the foundations associated with a moral expression. For instance, a moral sentiment might seem strongly related to authority but could also be linked to loyalty, leading to uncertainty about whether to label it as authority alone or both authority and loyalty.

The second source of ambiguity arises from the challenge of discerning the intended moral relevance from an author's language. For example, a social media post stating, "Everything that is going on with abortion these days is reprehensible," is evidently a morally charged statement. However, the specific foundation it pertains to is less clear. If the author is a secular liberal concerned with civil rights, it might be inferred that the statement relates to the fairness/cheating foundation due to concerns about women's reproductive rights. Conversely, if the author is a conservative Christian, the statement might reflect an anti-abortion sentiment associated with purity/degradation. Thus, the same expression can convey different moral sentiments, and competing interpretations can be challenging, if not impossible, to resolve systematically.

These ambiguities pose significant challenges for human annotators, who must find a balance between recognizing subtle moral sentiments and avoiding unwarranted assumptions about authorial intent. Excessive reliance on individual intuitions can lead to inconsistencies among coders, while overly literal interpretations can overlook the nuances of human language and morality. Therefore, a balance must be achieved between implicit coding, which involves inferences about authorial intent, and explicit coding, which focuses on the literal interpretation of the text.

Although achieving this balance perfectly is difficult, being mindful of these extremes can help limit coder biases. Since we typically lack access to the authors of the texts we analyze—and sometimes even the context of their discourse—we train annotators to primarily focus on explicit signals of moral sentiment and minimize inferences about authorial intent unless they are strongly defensible. This approach aims to reduce the risk of cultural biases introducing additional noise into the annotations. While our protocol strives to minimize annotator disagreement, we also caution against artificially reducing annotation variance.

When coding for MFT content, disagreements about which foundation is relevant are common. Even among expert coders, it is often unclear which perspective is correct. While some disagreements can be resolved through discussion, excessive resolution can artificially inflate intercoder reliability. Moral values are inherently subjective, and true accuracy of a code cannot be determined objectively. The closest approximation to an objective criterion is consensus among a constituency. As consensus is approached, the certainty that a phenomenon is strongly associated with a specific Moral Foundation increases. Low consensus among trained coders, therefore, is not merely a problem but an indication that the association might be subject to important boundary conditions, weak, or even illusory. Training coders to minimize disagreement does not change this reality but conceals it. Consequently, while coders need training, it should focus on establishing a shared understanding and heuristics for generating codes without fabricating agreement.